\pdfoutput=1
\documentclass[11pt]{article}

\usepackage{emnlp2021}

\usepackage{times}
\usepackage{latexsym}
\usepackage{balance}
\usepackage[T1]{fontenc}
\usepackage[utf8]{inputenc}

\usepackage{microtype}

\usepackage{times}
\usepackage{latexsym}
\usepackage{balance}

\newcommand{\OurDataset}{ShadowLink}

\usepackage{tikz}
\def\checkmark{\tikz\fill[scale=0.4](0,.35) -- (.25,0) -- (1,.7) -- (.25,.15) -- cycle;}

\usepackage{microtype}

\usepackage[disable]{todonotes}
\presetkeys{todonotes}{inline}{}
\usepackage{caption}
\usepackage{subcaption}
\usepackage{multirow}
\usepackage{booktabs}

\usepackage[inline]{enumitem}

\title{Robustness Evaluation of Entity Disambiguation Using Prior Probes:\\
the Case of Entity Overshadowing}
\author{Vera Provatorova,~ Svitlana Vakulenko,~ Samarth Bhargav,~ Evangelos Kanoulas \\
  University of Amsterdam, Amsterdam, The Netherlands \\
  \texttt{\{v.provatorova,  s.vakulenko, s.bhargav, e.kanoulas\}@uva.nl}
 }
\date{}

\begin{document}
\maketitle
\begin{abstract}
%  final versions of long papers will be given one additional page of content (up to 9 pages) so that reviewers’ comments can be taken into accoun
% \sv{Long papers may consist of up to 8 pages of content, plus unlimited pages for references}
Entity disambiguation (ED) is the last step of entity linking (EL), when candidate entities are reranked according to the context they appear in.
All datasets for training and evaluating models for EL consist of convenience samples, such as news articles and tweets, that propagate the prior probability bias of the entity distribution towards more frequently occurring entities.
% In this paper, we show that
It was previously shown that performance of EL systems on such datasets is overestimated, since it is possible to obtain higher accuracy scores by merely learning the prior.
% without the need to perform ED in most of the cases.
% , since the evaluation datasets follow the same distribution.
% Thereby, the systems may obtain high accuracy scores on such datasets by merely learning the prior without the need to perform ED in most of the cases, since the evaluation datasets follow the same distribution.
% From these considerations, we conjecture that the performance of the existing EL algorithms on the ED task is overestimated.
To provide a more adequate evaluation benchmark, we introduce the ShadowLink dataset, which includes 16K short text snippets annotated with entity mentions.
% resolved to Wikipedia pages. 
% , specifically designed to evaluate NED performance in a more systematic way
% \todo{revise here structure and content and number of samples}
% All text snippets are grouped in pairs, where each pair shares the same mention of two different entities anchored in one of the Wikipedia disambiguation pages.
We evaluate and report the performance of several popular EL systems on the ShadowLink benchmark. The results show a considerable difference in accuracy between common and uncommon ambiguous entities that require disambiguation, for all of the EL systems under evaluation, demonstrating the effects of prior probability bias and entity overshadowing.
\end{abstract}
\section{Introduction}
\label{sec:intro}
The task of entity linking (EL) refers to finding named entity mentions in unstructured documents and matching them with the corresponding entries in a structured knowledge graph \citep{milne2008learning,oliveira2021towards}.
This matching is usually done using the surface form of an entity, which is a text label assigned to an entity in the knowledge graph~\citep{van2020rel}.
Some mentions may have several possible matches: for example, ``Michael Jordan'' may refer either to a well-known scientist or the basketball player, since they share the same surface form.
Such mentions are ambiguous and require an additional step of entity disambiguation (ED), which is conditioned on the context in which the mentions appear in the text, to be linked correctly.
Following \citeauthor{van-erp-groth-2020-towards} (\citeyear{van-erp-groth-2020-towards}) we refer to a set of entities that share the same surface form as an \textit{entity space}.

\begin{figure}[h!]
     \centering
     \begin{subfigure}[b]{\columnwidth}
        %  \centering
        %  \includegraphics[width=0.75\textwidth]{lehmann.png}
        %  \includegraphics[width=\textwidth]{lehmann1-crop.pdf}
        \includegraphics[scale=0.9,width=\textwidth]{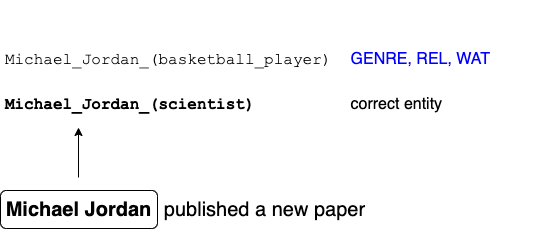}
         \caption{``Michael Jordan'' (scientist) is overshadowed by ``Michael Jordan'' (basketball player).}
         \label{fig:lehmann_1}
     \end{subfigure}
    %  \hfill
    %  \hspace{0.25cm}
     \begin{subfigure}[b]{\columnwidth}
         \centering
        \includegraphics[width=\textwidth]{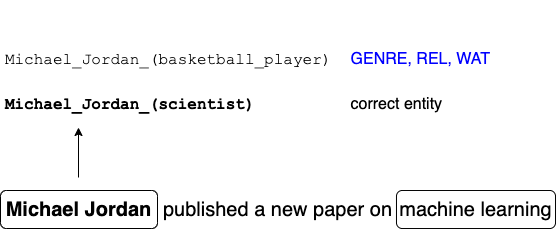}
         \caption{Even with more relevant context, overshadowing persists.}
         \label{fig:lehmann_2}
     \end{subfigure}
     \caption{An example of entity overshadowing. The correct entity is ranked lower by the EL systems (indicated in blue) than the more common one.}
\end{figure} 

% \sv{keep an eye on the margins the figure should not go beyond the text margin. To have the same size of the two figures make sure they are the same width, continue the sentence from new line, maybe remove the cyclist from the top figure if none of the systems predicts it in that case}

% \sv{chunk the following wall of text into passages, check topic sentences}
To decide which of the possible matches is the correct one, an ED algorithm typically relies on:
\begin{enumerate*}
\item contextual similarity, which is derived from the document in which the mention appears, indicating the \textit{relatedness} of the candidate entity to the document content, and
\item entity importance, which is the prior probability of encountering the candidate entity irrespective of the document content, indicating its \textit{commonness}~\citep{milne2008learning,ferragina2011fast,van2020rel}.
\end{enumerate*}
% \sv{what are the most common approaches to estimate the two, eg. in REL and TagME?}

% While commonness is based on the natural distribution of the data and therefore favours more popular entities, using relatedness is supposed to mitigate this effect by helping the system choose the entities that are more relevant for a particular document. \\
% However, it is currently not clear whether there is an optimal balance between using commonness and relatedness in the decisions of state-of-the-art ED models.

The standard datasets currently used for training and evaluating ED models, such as AIDA-CoNLL~\citep{hoffart2011robust} and WikiDisamb30~\citep{ferragina2011fast}, are collected by randomly sampling from common data sources, such as news articles and tweets.
Therefore, they are expected to mirror the probability distribution with which the entities occur, thereby favouring more frequent entities (head entities)~\citep{ilievski-etal-2018-systematic}.
From these considerations, we conjecture that the performance of existing EL algorithms on the ED task is overestimated.
We set out to explore this effect in more detail by introducing a new dataset for ED evaluation, in which the entity distribution differs from the one typically used for training ED algorithms.
% since the evaluation data  follows the same distribution as their training data biased towads more common entities.

% Such datasets are not sufficient for evaluating the ability of the models to use context-based relatedness for disambiguation: if a model is biased towards more popular entities and ignores less popular ones even when they are more relevant, it may still perform well enough to get high evaluation scores. \\

%Poor performance on the long-tail entities was previously reported by \citeauthor{ilievski-etal-2018-systematic} (\citeyear{ilievski-etal-2018-systematic}).
We perform a systematic study focusing on a particular phenomenon we refer to as \textit{entity overshadowing}.
Specifically, we define an entity $e_1$ as overshadowing an entity $e_2$ if two conditions are met: \begin{enumerate*}
    \item $e_1$ and $e_2$ belong to the same entity space $S$, i.e., share the same surface form and, therefore, can be confused with each other outside of the local context;
    \item $e_1$ is more common than $e_2$ in some corresponding background corpus (e.g. the Web), i.e., it has a higher prior probability $P(e_1) > P(e_2)$.
\end{enumerate*}

For example, $e_1$ =  ``Michael Jordan'' (basketball player) overshadows $e_2$ = ``Michael Jordan'' (scientist) because $P(e_1) > P(e_2)$ in a typical dataset sampled from the Web.
We use an unambiguous text sample that contains this mention to evaluate three popular state-of-the-art EL systems, GENRE~\citep{de2020autoregressive}, REL~\citep{van2020rel}, and WAT~\citep{piccinno2014tagme}, and empirically verify that the overshadowing effect that we hypothesized, indeed, takes place (see Fig.~\ref{fig:lehmann_1}).
Even when more information is added to the local context, including the directly related entities that were correctly recognised by the system (``machine learning''), the ED components still fail to recognise the overshadowed entity (see Fig.~\ref{fig:lehmann_2}).

The concept of overshadowed entities introduced in this paper is related to long-tail entities \citep{ilievski-etal-2018-systematic}.
However, these two concepts are distinct: a long-tail entity may be unambiguous and therefore not overshadowed, while an overshadowed entity may still be too popular to be considered a long-tail one.

% The entities in the long-tail are either infrequent

% \todo{mention relation to prior work on long tail entities by Philipp and show what we do differently here}

% To explore this problem in more detail, we introduce the concept of \textit{entity overshadowing}.

% \begin{tabular}{cc}
%      \includegraphics[width=0.23\textwidth]{fig1a_lehmann.png}& \includegraphics[width=0.23\textwidth]{fig1b_lehmann.png}  
% \end{tabular}
%
% 
% To the best of our knowledge, none of the existing datasets are suitable for detecting entity overshadowing phenomena. Thus
To systematically evaluate the phenomenon of entity overshadowing that we have identified, we introduce a new dataset, called \OurDataset{}. \footnote{ShadowLink dataset can be downloaded at \url{https://huggingface.co/datasets/vera-pro/ShadowLink}}.
% \todo{...something about the dataset: briefly how was constructed, how big, what are the main parts}
\OurDataset{} contains groups of entities that belong to the same entity space.
Following \citeauthor{van-erp-groth-2020-towards} (\citeyear{van-erp-groth-2020-towards}), we use Wikipedia disambiguation pages to collect entity spaces.
Disambiguation pages group entities that often share the same surface form and may be confused with each other.
We then follow the links in the Wikipedia disambiguation pages to the individual (entity) Wikipedia pages to extract text snippets in which each of the ambiguous entities occur.

Note that we do not extract the text from these Wikipedia pages directly, since pre-trained language models such as BERT (typically used in state-of-the-art ED systems) also use Wikipedia as a training corpus, and can learn certain biases as well.
Instead, we parse external web pages that are often linked at the end of a Wikipedia page as references.
This data collection approach helps us to minimise the possible overlap between the test and training corpus. 

Thereby, every entity in \OurDataset{} is annotated with a link to at least one web page in which the entity is mentioned.
We then proceed to extract all text snippets in which the corresponding entity mention appears on the page.
An extracted text snippet typically consists of the sentence in which the mention occurs.%, one sentence before and one after.
% of average length of 25 words %\sv{? sentences} 

Next, we use \OurDataset{} to answer the following research questions:

% \textbf{RQ1:} How can we evaluate the impact of (local) context on ED performance?\\
% \todo{make sure that the RQs are consistent with experiments/results sections}
% \noindent
% \textbf{RQ2:} Are EL predictions biased towards more common entities? How can we measure commonness and is it predictive of the EL performance?\\
%\sv{do we need all 4 RQs? can we group them to make more readable? or embed in the intro text instead of listing?}
\textbf{RQ1:} How well can existing ED systems recognise overshadowed entities?

\textbf{RQ2:} How does performance on overshadowed entities compare to long-tail entities?

\textbf{RQ3:} Are ED predictions biased and how can we measure this bias?
% \textbf{RQ4:} How does human performance on the overshadowed entities compare with the automated results? 
% \todo{add other RQs}
% how well do the models perform on our data in general, and (b) to which extent they are able to use the context (relatedness) to make the prediction instead of relying on the commonness prior.

% \todo{brief description of our experiments}
% \todo{brief description of results – will be added later}
% \sv{so are we evaluating EL or ED?}

Our contribution is twofold:
\begin{enumerate*}
\item a new dataset for evaluating entity disambiguation performance of EL systems specifically focused on overshadowed entities, and
\item an evaluation of current state-of-the-art algorithms on this dataset, which empirically demonstrates that we correctly identified the type of samples that remain challenging and provide an important direction for future work. 
\end{enumerate*}
% \todo{the rest of the paper is structured as follows (if need be)}
% \input{31_dataset_construction.tex}
% \input{32_dataset_structure.tex}
\section{The ShadowLink Dataset}
% \todo{what is the structure of the dataset?}
This section describes the ShadowLink dataset: its construction process, structure, and statistics.
% Dataset description starts here
% \sv{what is this section about? what are the subsections and how they relate}
\subsection{Dataset construction} %structure goes here as well
% \sv{how was the dataset constructed?}
The process of dataset construction consists of 3 steps: (1) collecting entities, (2) retrieving context examples for each entity, and (3) filtering the data based on the validity requirements detailed below. 
% \sv{We used Wikipedia disambiguation pages to provide examples of entity spaces...}
% \resizebox{\columnwidth}{!}{
% \sv{the font should be bigger to make the figure more readable? reduce spaces between or shorten the text} UPD: done

\begin{figure}[h!]
    % \centering
    \includegraphics[width=\columnwidth]{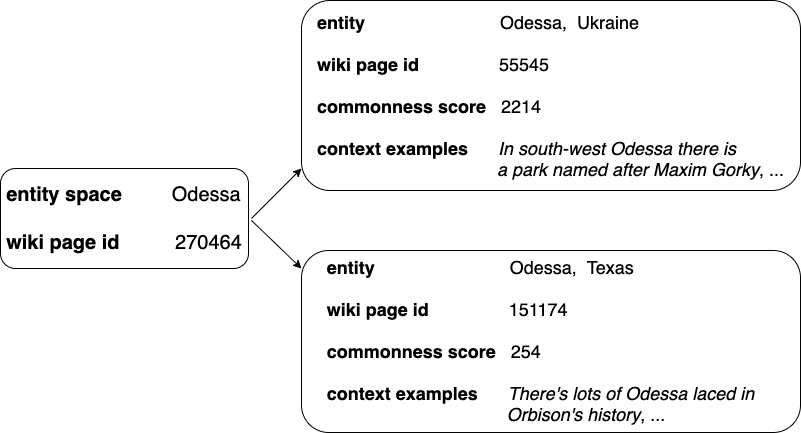}
    \caption{Structure of the ShadowLink dataset}
    \label{fig:dataset_diagram}
\end{figure}
% }
%

\textbf{Collecting entities.} Similar to \citet{van-erp-groth-2020-towards}, we use Wikipedia disambiguation pages to represent entity spaces. We retrieve a set of all Wikipedia disambiguation pages and filter it on the following criteria:
\begin{enumerate}
    \item For each disambiguation page (DP), we only include candidate entity pages with names containing the title of the DP as a substring. This step is required to exclude synonyms and redirects. %For example, if we have a DP with a title ''Petkus'', then its candidate entity page ''Andrius Petkus'' is included, while ''Red Guardian'' is not.
    % \item If a candidate page link is not active (deleted or not created yet), then it is excluded.
    \item If at least two candidate pages for the same DP match the criterion described above, then the DP and all its matching candidates are included as a new entity space.% (see Figure~\ref{fig:dataset_diagram} for an illustration of an entity space).
\end{enumerate}
% we probably need a better name than "candidate page" here
During the first stage of the data collection, 170K out of 316.5K Wikipedia disambiguation pages matched the filtering criteria described above.

\textbf{Filtering pages by year.} To make sure that all pre-trained EL systems we evaluate in our experiments can potentially recognise all of the entities in the dataset, we also exclude pages that are more recent than the Wikipedia dumps used by these systems during training. The oldest dump used by a system in our experiments was the 2016 Wikipedia dump over which TagMe was trained, i.e we excluded all the pages that were created after 2016.
% \sv{This resulted in x pages}
% \todo{how many pages were filtered out by this criteria?}

\textbf{Collecting context examples.} To retrieve context examples for each entity, we follow the external links extracted from the references section of the corresponding Wikipedia page and parse them to extract the text snippets which contain the entity mention. Then, every target entity mention is replaced with its corresponding entity space name, yielding an ambiguous entity mention. For example, if we have entities "John Smith" and "Paul Smith" that both belong to the entity space "Smith", then the mentions of both names will be replaced with "Smith". Looking for an entity name and replacing it with the corresponding entity space name (instead of looking for the entity space name in the first place) allowed us to make sure that the text snippets refer to the correct entity. Using this method, however, significantly reduced the number of retrieved snippets, as many of the entity mentions in natural texts do not include the full titles of the entities.

To extract the text snippets, we used a simple greedy algorithm that starts with the mention boundaries and tries to include more text, expanding the boundaries to the left and to the right, until it either covers one sentence on each side, or reaches the end (or beginning) of the document text. Our decision was to use relatively short spans similar to other popular ED benchmarks: WikiDisamb30 \citep{ferragina2011fast} and KORE50 \citep{hoffart2011robust}. Our manual evaluation confirmed that these spans provide sufficient context for entity disambiguation. We also release the full-text of all web pages as part of our dataset, making the context of different lengths available for future experiments. \todo{updated info about choosing context - check if it looks alright}
% Thus, the intended length of a snippet is 3 sentences; however, due to the noisy nature of web pages this often resulted in a single sentence, the one that contained the mention of the entity.

% Previous datasets on ED use various snippet sizes. For instance, WikiDisamb30 \cite{ferragina2011fast} and KORE50~\cite{DBLP:conf/cikm/HoffartSNTW12} use single sentences containing the entity mention, while AIDA-CoNLL~\citep{hoffart2011robust} uses entire articles.

%The snippet size was chosen as a heuristic: long enough to contain meaningful context, and short enough to avoid too much noise\footnote{}. 
% \todo{it says average is 2 in the introduction}
% \todo{why? give example to provide an intution} \upd{added example}
%
% \todo{how is the snippet extracted? what is the length of the snippet? why? is it similar to any of the other datasets?}
% \todo{explain the intution behind popularity-commonness choice to estimate priors based on the ED algorithms}

\textbf{Commonness score.} We estimate the commonness (popularity) of an entity as the number of links pointing to the entity page from other Wikipedia pages, that is, the in-degree of the entity page in the web graph of Wikipedia hyperlinks.
Intuitively, this is proportional to the probability of encountering this entity when sampling a page at random.
To obtain this metric for all the entities in the dataset, we use the Backlinks MediaWiki API\footnote{\url{https://www.mediawiki.org/wiki/API:Backlinks}}. 
% We compute popularity of an entity page $p$ as the number of links pointing to $p$ in the Wikipedia dump of 20.11.2020.\\
% we assume that popularity hasn't changed much since 2016
% \sv{why do we need popularity measure? where is this defined and motivated: popularity, top, overhadowed? can we highlight these pieces in the text to help reader to locate main concepts?}

\textbf{Quality assurance.} We conduct manual evaluation to assess the quality of the dataset and provide the upper bound performance for the ED task. The details of the setup and the results are discussed in Section~\ref{sec:manual}.
\subsection{Dataset structure and statistics}
The ShadowLink dataset consists of 4 subsets: \textit{Top}, \textit{Shadow}, \textit{Neutral} and \textit{Tail}. The Top, Shadow and Neutral subsets are linked to each other through the shared entity spaces. On the other hand, the Tail subset, which contains (typically unambiguous) long-tail entities, is not connected to the other three through the same entity spaces. Nevertheless, it is collected in a similar way as the other three subsets.
% \sv{how many samples in total? how many in the splits?}
% \textbf{Test split.}
% \todo{Structure: subsets Top/Overshadowed/Neutral/Tail}
% \todo{describe the structure on the high level using the example in the figure 2: DP page, external pages, context samples}

\textbf{Top and Shadow subsets.}
The structure of the Top and Shadow subsets is shown in Figure~\ref{fig:dataset_diagram}.
Every entity $e$ belongs to an entity space $S_m$, derived from the Wikipedia disambiguation pages, where $m$ is an ambigous mention that may refer to any of the entities in $S_m$.
Every $S_m$ contains at least two entities: one $e_{top}$ and one or more $e_{shadow}$ entities.
% , which were selected in such a way that $C(e_{top})~\geq~C(e_{shadow})~+~10$.
Every entity $e \in S_m$ is annotated with a link to the corresponding Wikipedia page and provided with context examples.
A context example is a text snippet extracted from one of the external pages which contains the mention $m$ , with a length of 25 words on average.
% of the entity.
% The surface form of $m$ coincides with $S$.
% For example, if $S$ = ``Smith'', $e_{top}$=``John Smith'' and $e_{shadow}$=``Balthazar Smith'', then both entities are mentioned as ``Smith'' in their corresponding context examples.
% \todo{explain $m$ and give example}
% \todo{do we have more than one external pages, more than one context example?}
% \textbf{ShadowLink.}
% \todo{what about the new long-tail subset? how many samples are there in total? \#entity spaces/\#entities/\#context pages/\#text snippets}
% \upd{only one context page per entity was included (due to time constraints: found a page that contains good examples -> stop looking further); and one example per page (but more examples are available)}
% \todo{explain top and overshadowed subsets.}
% \todo{Why? How?}

\textbf{Neutral subset.}
To quantify the strength of the prior of each ED system, we synthetically generate data points for which the context around an entity mention is not useful for disambiguating that mention. To do that we use 7 hand-crafted templates. An example of such a template is the following: "It was the scarcity that fueled our creativity. This reminded me of $m$ today." For each entity space, we generated 7 random contexts.
% We then report median and variance.}
% \sv{pick a context example that is more similar to context examples we have in the dataset, e.g., taking the median legth? It should probably be more like 2 sentences instead.}
%When an entity mention is used in a neutral context, there is no relevance signal and the system is likely to rely on its priors to increase its chances to make the correct prediction without recognising the context.
%Thus, by using such samples, we try to find out the systems' priors, which may differ across the systems.
% that dectate the ``default predictions'' 

\textbf{Tail subset.}
To evaluate the performance of ED systems on long-tail but typically not overshadowed entities, we collect an additional set of entities by randomly sampling Wikipedia pages that have a low commonness score ($<=56$ backlinks)\footnote{This threshold is equal to the median number of backlinks in the \textit{Shadow} subset.}. 
%The commonness threshold was selected as the median of the commonness measured for the overshadowed entities.
% It is designed to match the size of the main subset.
% The dataset with long-tail entities (TailLink) matches the size and structure of ShadowLink: it contains 904 entities provided with context examples. The entities were randomly selected from Wikipedia pages and filtered by applying inclusion criteria. To be included into TailLink, an entity $e$ must match the following criteria:
% \begin{enumerate}
%     \item Popularity of $e$ does not exceed the threshold of 56 backlinks. The threshold was obtained as the popularity median of the shadow subset in ShadowLink
%     % \item The entity $e$ is not ambiguous % not sure about this one: it was a bit of a heuristic (most entities are not ambiguous -> include everything and hope for the best). will probably check if they are really unambiguous
%     \item The Wikipedia page of $e$ was created not later than in 2016
% \end{enumerate}

%  as for the main subset: filtering by year, parsing external pages listed as references, and extracting text snippets with entity mentions
Context examples for these pages were collected in the same manner as described above. The resulting dataset matches the size and structure of other ShadowLink subsets, containing 904 entities.

The sampling process used to collect this subset follows the existing definition of long-tail entities\citep{ilievski-etal-2018-systematic}, and is controlled for popularity but not for ambiguity. %, and therefore the collected entities may or may not be ambiguous. 
The Tail subset serves as a control group for the experiments conducted in our study, showing that the concept of entity overshadowing differs from the previously studied long-tail entity phenomena.

\begin{figure}[h!]
    \centering
    \includegraphics[width=0.8\columnwidth, height=5cm,keepaspectratio]{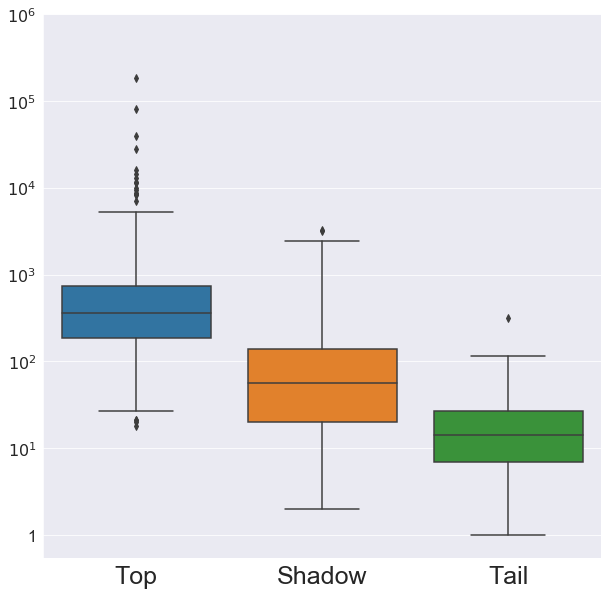}
    \caption{Distribution of the commonness score on the three subsets of \OurDataset{}.}
    \label{fig:commonness_distribution}
\end{figure}

\textbf{ShadowLink statistics.} The dataset statistics across all the subsets are summarised in Table~\ref{tab:dataset_stats}.
Note that the \textit{Top}, \textit{Shadow} and \textit{Neutral} subsets are grouped around the same entity spaces, while the \textit{Tail} subset is constructed by sampling the same number of non-ambiguous entities.  Every entity space contains at least 2 entities, with the mean number of entities per space being 2.63, median 2, and maximum 10.
\todo{added statistics on \# entities per space as promised to reviewer 2, does it look alright?}
Figure~\ref{fig:commonness_distribution} shows the distribution of commonness in the three subsets: Top, Shadow and Tail.

\begin{table*}[]
\centering
% \small
\begin{tabular}{lccccc}
\hline 
% per Snippet
\textbf{Subset} & \textbf{\# Entity Spaces} & \textbf{\# Entities} & \textbf{\# Text Snippets} & \textbf{Avg. \# Words} & \textbf{Avg. \# Sentences} \\ 
\hline
Top & 904 & 904 & 2K & 29.25 & 1.11 \\
Shadow & 904 & 1.5K & 6K & 28.97 & 1.11 \\
Neutral & 904 & - & 6K & 14.83 & 1.87 \\
Tail & - & 904 & 2K & 28.94 & 1.10 \\ \hline
\end{tabular}
\caption{\label{tab:dataset_stats} Dataset statistics across all the subsets of \OurDataset{}. The average number of words and sentences were calculated per text snippet extracted from the corresponding web page.}
\end{table*}

For the experiments we used a smaller subset of ShadowLink, with only one randomly selected shadow entity per entity space and one text snippet per entity. Thus, every subset contained 904 entities, with the total size of 9K text snippets. The rest of the data is left out as a training set and can be used in future experiments.
% \todo{is it possible to make the labels font larger? replace top commonnness with Top etc.}
% done
\section{Manual Evaluation}
\label{sec:manual}
We perform manual evaluation of a random sample from \OurDataset{} to assess its quality, with the goal of ensuring that the extracted text snippets provide context sufficient for disambiguation.
Human performance also sets the skyline for automated approaches on this dataset.
% Therefore, we are interested in the accuracy that human annotators achieve on \OurDataset{} (RQ4).
In the following subsections, we describe the evaluation setup and the results of the manual evaluation.
% \todo{explain why and what and how and structure of the subsections}
% \subsection{Manual Evaluation}
% \label{sec:manual}
\subsection{Manual evaluation setup}
% \sv{We conducted a manual evaluation to assess the human performance and assess the quality of the dataset. Two annotators examined n randomly selected samples from our test set independently and ... what were they instructed to do? what is input/output?}
% We also included a human annotation baseline, which consisted of 150(?) randomly selected dataset entries.
We conduct a manual evaluation to assess the quality of the dataset and evaluate how well human annotators can disambiguate overshadowed entities. A sample of 91 randomly selected dataset entries was presented to two annotators, who examined the entries independently.
%\todo{what are the proportions? was the set shuffled?}
For each entry, the annotators were presented with a text snippet containing an ambiguous entity mention $m$, and two entities, \textit{Top} and \textit{Shadow}, from the same entity space $S_m$, where one of the two entities was the correct answer.
% which can be linked to either \textit{top} or \textit{overshadowed} entity. 
The annotators were instructed to either indicate the correct entity or mark the text snippet as ambiguous, which indicates that the provided context is not sufficient for the disambiguation decision to be made. Note, however, that the commonness scores were not displayed to the annotators.

\subsection{Results of the manual evaluation}
% To find out whether human annotators can disambiguate overshadowed entities better than algorithms, we conducted manual evaluation as described in \ref{sec:manual}. For each example, the annotators could either choose one of the two answers, or indicate that the context does not provide enough information. \\
%
% \textbf{Evaluating the dataset.} The results of the first step of manual evaluation are presented in Table~\ref{tbl:manual-all}, which shows how many examples were labelled by the annotators as not clear enough, and how many of the remaining entities were disambiguated correctly.
% , both annotators disambiguated 82 entities correctly (hence, 85\% accuracy score). The first annotator disambiguated 2 entities incorrectly and marked 12 samples as ambiguous given the context. The second annotator gave 1 incorrect answer and marked 13 samples as ambiguous. The wrong answers of the annotators did not overlap.
% Therefore, we conclude that they are due to the errors made by the annotators rather than by the automated extractor.\\
% \textbf{Interannotator agreement.} 
\begin{table*}[h!]
\centering
\begin{tabular}{@{}lcc@{}}
\toprule
            & \textbf{Shadow}            & \textbf{Top} \\ \midrule
            & \multicolumn{1}{c|}{P~=~R~=~F} & P~=~R~=~F        \\
Annotator 1 & \multicolumn{1}{c|}{0.973} & 0.973        \\
Annotator 2 & \multicolumn{1}{c|}{0.950} & 0.919        \\
Average     & \multicolumn{1}{c|}{0.963} & 0.946        \\ \bottomrule
\end{tabular}
\caption{\label{tbl:human} Results of the manual annotations.}
\end{table*}

% \sv{we don't need to show F P R if they are the same why are they the same? is it accuracy?}

We used Cohen's kappa coefficient to evaluate the inter-annotator agreement~\citep{bobicev2017inter} on all entries reviewed by the annotators. The value of the coefficient is 0.845, indicative of strong agreement.
Next, we discarded the samples labelled as ambiguous by at least one of the annotators. The resulting dataset included 77 entries out of 91, which shows that 85\% of the context examples were sufficient for making ED decisions. These unambiguous entries were split into two subsets, resulting in the 37 top-entities and 40 shadow-entities. We then discarded 3 randomly selected shadow-entities to achieve the same size of the two subsets, and used these subsets to evaluate the performance of manual ED for the top- and shadow-entities separately.
%By design of the experiment, there are no false negative answers, and therefore precision is the same as accuracy while recall is 1.
% \todo{add the average result to the table in the results as a skyline for automated approaches}
% We report this result in Table~\ref{tab:sample_results} as a skyline for the automated performance on the ShadowLink dataset.
The averaged F-score of the two annotators is 0.95 on the top-entities and 0.96 on the shadow-entities.  The detailed results of the evaluation are shown in Table~\ref{tbl:human}. 

The results of manual evaluation show that (1)~a majority of samples (85\%) in \OurDataset{} are suitable for ED evaluation, i.e., automatically extracted snippets provide sufficient context for correct disambiguation; (2) human annotators can correctly disambiguate entities regardless of their commonness. Therefore, the performance of an automatic system that only depends on context is only bound by the 15\% of the cases for which the context is not helpful. This bound can be further elevated if longer contexts are considered. Experiments on longer contexts are possible using the \OurDataset{} dataset\footnote{We have crawled the full articles, and will be released as part of the \OurDataset{} datset.} but we leave it for future work.

In the next section, we report and analyse the results produced by state-of-the-art systems on \OurDataset{}.
% For additional quality assurance, we also measure the performance of all baselines on the manually controlled sample. The results are consistent with the results obtained on the full ShadowLink dataset. The detailed report can be found in the Appendix~\ref{sec:appendix}.
% \todo{Add P R F table for humans?}
% The results are presented in Table~\ref{tbl:manual-top-shadow}.
% \begin{table}[]
% \begin{tabular}{rrr}
%                      & \textbf{top sample} & \textbf{shadow sample} \\
% \textbf{Annotator 1} & 0.9756              & 0.9348                 \\
% \textbf{Annotator 2} & 0.9512              & 0.9348                
% \end{tabular}
% \caption{\label{tbl:manual-top-shadow} Manual evaluation results on top and shadow subsamples}
% \end{table}
%
% \todo{interannotator agreement here: 0.9545}
% \todo{how was it calculated?}
% Context window: 2-3 sentences? \\
% \section{Evaluation Setup}
\section{Benchmark Experiments}
\label{sec:benchmark}

% \sv{the paragraph is too long. split into two?} UPD: done

% [it's a "free writing" draft, will be improved] \\
In this section, we describe the benchmark experiments designed to evaluate the baseline systems' performance on the ShadowLink dataset.
% \todo{how we do it? considering performance for different subsets}
For these experiments, we created a subset of the original dataset by sampling only one of the shadow entities at random to make the number of \textit{Top} and \textit{Shadow} equal.
Note that in our task setup the model's predictions are not restricted to the top versus shadow entity binary decision.
The model can predict any entity from the same or different entity space.
We describe the experimental setup in Section~\ref{sec:eval_setup}, report the benchmarking results and analyse them in more detail in Section~\ref{sec:benchmark_res}.
\subsection{Evaluation setup}
\label{sec:eval_setup}
% \todo{where do we mention the setup, input text for EL+ED or define correct spans}
% [REL, Wat, TagMe, BLINK, Bootleg (?) + models from Gerbil + human baseline]
% The evaluation process consisted of two steps. First, we evaluated the performance of eight entity linking systems on the ShadowLink dataset.
To answer the first two research questions (RQ1 \& RQ2), we compare the performance of eight entity linking systems on the ShadowLink dataset.
We used the GERBIL framework \citep{roder2018gerbil} for six of the baselines (AGDISTIS/MAG, AIDA, DBpedia Spotlight, FOX, TagMe 2 and WAT)\footnote{These six systems were the ones available on GERBIL at the time of our experiments.} under the D2KB experimental setup\footnote{In the D2KB setup, the systems are provided with correct mention boundaries to evaluate the disambiguation step of entity linking.}.
We also performed an evaluation with the same setup using GENRE and REL, two novel state-of-the-art systems not available in GERBIL. 
We used micro-averaged precision, recall and F-score as evaluation metrics.
% to compare the systems.
% \todo{maybe AIDA here if we make it + maybe refine the selection arguments}
% We then selected three systems with the best results and carried out a detailed analysis of their predictions to evaluate the impact of prior bias on disambiguating overshadowed entities.

To answer the last research question (RQ3), we want to verify whether the baseline systems utilise context or simply rely on their priors to make the predictions.
To this end, we compare the predictions made on the \textit{Top}, \textit{Shadow} and \textit{Neutral} subsets.
% As a reminder
% \sv{what about Neutral size? is it the same size as Top=Shadow or Top+Shadow?}
We used the predictions made on the \textit{Neutral} subset as an indication of priors.
That is, for each entity space, we generate context for the \textit{Neutral} subset by using the same 7 random sentences as templates.
% as an input to the evaluated systems.
The context was generated as neutral, i.e., it is not useful for the disambiguation task by design.
Therefore, we considered the predictions for such neutral contexts to exhibit the default priors of an EL system for the given entity space.
We can then compare these prediction to the predictions on the original examples from the \textit{Top} and \textit{Shadow} subsets.
% to evaluate whether the system changes its decisions.% when presented with different context.
% \sv{remind how neutral examples were created. are they created for the same entities as Top and Shadow samples?}
If the entity predicted for non-neutral context differs from the prediction made for the neutral context, we consider that the model updated its default prediction (prior) based on the local context.
% Otherwise, the local context was not used for disambiguation and the system relied on the prior.
% \todo{added a clarification about ED being non-binary (for reviewer 1), does it look alright?}
We performed this type of analysis to examine the predictions of the best-performing systems in our experiments: REL, GENRE, AIDA and WAT.

\subsection{Benchmark Results}
\label{sec:benchmark_res}
% \section{Benchmark Experiments}
% \todo{briefly describe the content as a whole what are we going to present here, repeat and answer the RQs}
This section presents the results of our experiments  and summarizes the answers to the research questions introduced in Section~\ref{sec:intro}.

\textbf{RQ1:} \textit{How well can existing ED systems recognise overshadowed entities?}

Table~\ref{tbl:res-all} shows the evaluation results across the subsets of ShadowLink.
All systems achieve the lowest scores on the \textit{Shadow} subset, with the maximum F-score of 0.35 achieved by AIDA.
While REL and GENRE ourperform WAT on several existing datasets~\citep{van2020rel, de2020autoregressive}, their results on ShadowLink are considerably lower than the results of WAT.
The difference in the results on \textit{Top} and \textit{Shadow} entities indicates that EL predictions are biased towards more common entities.

\begin{table*}[h!]
\centering
\begin{tabular}{@{}l ccc ccc ccc@{}}
\toprule
\multicolumn{1}{c}{\multirow{2}{*}{\textbf{Baseline}}} & \multicolumn{3}{c}{\textbf{Shadow}} & \multicolumn{3}{c}{\textbf{Top}} & \multicolumn{3}{c}{\textbf{Tail}} \\ \cmidrule(l){2-10} 
\multicolumn{1}{c}{} & P & R & F & P & R & F & P & R & F \\ \midrule
AGDISTIS/MAG~\citep{usbeck2014agdistis} & 0.14 & 0.14 & \multicolumn{1}{l|}{0.14} & 0.25 & 0.25 & \multicolumn{1}{l|}{0.25} & 0.79 & 0.79 & 0.79 \\
AIDA~\citep{aida} & \textbf{0.40} & \textbf{0.31} & \multicolumn{1}{l|}{\textbf{0.35}} & 0.62 & 0.50 & \multicolumn{1}{l|}{0.56} & 0.92 & 0.53 & 0.67 \\
DBpedia Spotlight~\citep{DBLP:conf/i-semantics/MendesJGB11} & 0.16 & 0.10 & \multicolumn{1}{l|}{0.12} & 0.41 & 0.25 & \multicolumn{1}{l|}{0.31} & \textbf{0.97} & 0.11 & 0.19 \\
FOX~\citep{speck2014named} & 0.17 & 0.07 & \multicolumn{1}{l|}{0.10} & 0.29 & 0.14 & \multicolumn{1}{l|}{0.19} & 0.82 & 0.35 & 0.49 \\
TagMe~2~\citep{DBLP:conf/cikm/FerraginaS10} & 0.34 & 0.25 & \multicolumn{1}{l|}{0.29} & 0.69 & 0.49 & \multicolumn{1}{l|}{\textbf{0.57}} & 0.95 & 0.74 & 0.83 \\
WAT~\citep{piccinno2014tagme} & \textbf{0.40} & 0.19 & \multicolumn{1}{l|}{0.26} & \textbf{0.72} & 0.39 & \multicolumn{1}{l|}{0.51} & 0.95 & 0.49 & 0.65 \\
GENRE~\citep{de2020autoregressive} & 0.26 & 0.26 & \multicolumn{1}{l|}{0.26} & 0.42 & 0.42 & \multicolumn{1}{l|}{0.42} & 0.93 & \textbf{0.93} & \textbf{0.93} \\
REL~\citep{van2020rel} & 0.21 & 0.21 & \multicolumn{1}{l|}{0.21} & 0.54 & \textbf{0.54} & \multicolumn{1}{l|}{0.54} & 0.91 & 0.91 & 0.91 \\ \bottomrule
\end{tabular}
\caption{\label{tbl:res-all} Benchmark evaluation results on the ShadowLink subsets.}
\end{table*}

\textbf{RQ2:} \textit{How does the performance on overshadowed entities compare to long-tail entities?}

All systems show the highest precision on the \textit{Tail} subset, i.e., they achieve much higher performance on the less ambiguous long-tail entities, compared to both top and overshadowed entities in ShadowLink.
These results indicate that the main challenge in ED is the combination of ambiguity and uncommonness, while uncommon but non-ambiguous entities are relatively easy to resolve.

These findings are also consistent with \citeauthor{ilievski-etal-2018-systematic}~(\citeyear{ilievski-etal-2018-systematic}), who suggest that rare and ambiguous entities constitute the hardest cases for the EL task. In this study, we showed that such overshadowed entities indeed consititute a major challenge for the state-of-the-art systems and that \OurDataset{} provides a suitable benchmark for their evaluation.
% non-ambiguity or commonness alone do not suffice to characterize the level of difficulty.
% \todo{so what?}
% \sv{how can we explain this dependency?}}
% The results on the ShadowLink, however, are significantly lower, which indicates that high ambiguity is a bigger obstacle for ED than low commonness.
% \textbf{RQ4:} How does the performance on overshadowed entities compare to the long-tail but unambiguous entities?
\begin{table*}[]
\begin{tabular}{@{}l rrrrr rrrrr @{}}
\toprule
\multirow{3}{*}{} & \multicolumn{5}{c}{\textbf{Top}} & \multicolumn{5}{c}{\textbf{Shadow}} \\ \cmidrule(l){2-11} 
 & \multicolumn{2}{c}{pred = prior} & \multicolumn{2}{c}{pred $\neq$ prior} & \multicolumn{1}{c|}{\multirow{2}{*}{NIL}} & \multicolumn{2}{c}{pred = prior} & \multicolumn{2}{c}{pred $\neq$ prior} & \multicolumn{1}{c}{\multirow{2}{*}{NIL}} \\
 & correct & wrong & \multicolumn{1}{c}{correct} & \multicolumn{1}{c}{wrong} & \multicolumn{1}{c|}{} & \multicolumn{1}{c}{correct} & \multicolumn{1}{c}{wrong} & \multicolumn{1}{c}{correct} & \multicolumn{1}{c}{wrong} & \multicolumn{1}{c}{} \\ \hline
AIDA & 15.5 & 12.9 & 35.0 & 28.0 & \multicolumn{1}{r|}{8.6} & 4.8 & 20.0 & 28.2 & 39.3 & 7.9 \\
WAT & 32.1 & 11.7 & 16.4 & 12.4 & \multicolumn{1}{r|}{27.4} & 4.2 & 31.8 & 20.9 & 15.6 & 27.5 \\
GENRE & 15.6 & 32.4 & 26.7 & 26.3 & \multicolumn{1}{r|}{0.0} & 2.2 & 43.6 & 23.9 & 30.3 & 0.0 \\
REL & 32.3 & 24.8 & 21.4 & 21.0 &\multicolumn{1}{r|}{0.6} & 7.7 & 50.9 & 12.9 & 27.9 & 0.6 \\ \hline
\end{tabular}
\caption{\label{tbl:big-results-table} Error analysis, which shows the percentage of errors and correct predictions that either coincide (pred=prior) or differ (pred$\neq$prior) from the predictions made for the neutral contexts, which we consider as predictions with the highest prior probability.}
\end{table*}

% To further explore possible reasons behind the differences in the results, we turn to our next research question.
% This finding indicates that human annotators
% do not rely on the  can use local context more efficiently than algorithms do, which allows them to achieve high accuracy on shadow entities.
% \sv{make the font bigger in the figures}

\begin{figure}[ht!]
    % \centering
    \includegraphics[width=\columnwidth, height=4.5cm,keepaspectratio]{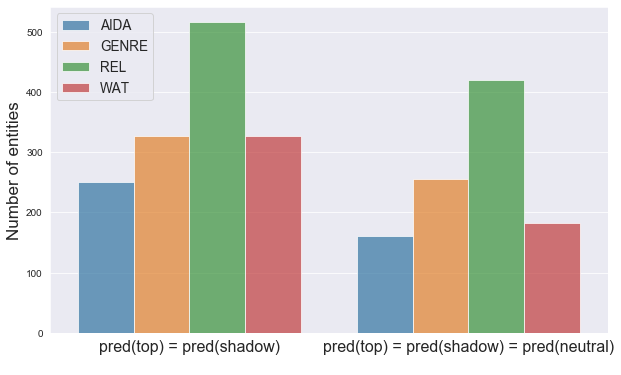}
    \caption{The degree of overshadowing (left) and prior bias (right) for each of the EL systems.} 
    % \sv{number of entities not answers we should stay consistent with terminology. can't see the numbers can be also percentage instead} upd: fixed
    \label{fig:prior_bias}
\end{figure}
%
% \textbf{Impact of priors.}
% \subsection{Detailed analysis of the ED predictions}
% \label{sect:priors}

\textbf{RQ3:} \textit{Are ED predictions biased and how can we measure this?}
% \sv{ED or EL? stick to one and check everywhere is consistent}

Our experiments show that all systems under evaluation are often insensitive to the context change, i.e., the systems are actually unable to exploit local context for entity disambiguation but solely rely on their priors learned from the data.
The error analysis results presented in Table~\ref{tbl:big-results-table} indicate that the majority of correct answers on the \textit{Top} dataset coincide with the predictions observed on the \textit{Neutral} subset.
% , and hence with the priors we identified.
On the \textit{Shadow} subset, opposite is the case: most of the errors are due to priors, and most of the correct predictions differ from them. 
% One exception is the AIDA system, hence the reason of the better performance observed in Table~\ref{tbl:res-all}.
%From this, we conclude that our commonness scores are a good proxy for the priors across all systems used on our evaluation.
% and 
% This indicates that the impact of commonness outweighs the impact of relatedness, preventing the systems from finding the correct answer despite the presence of context information.

Figure~\ref{fig:prior_bias} shows the number of cases in which overshadowing occurs for each of the systems, i.e., when the model's prediction remains the same for both \textit{Top} and \textit{Shadow} mentions.
We see that this effect correlates with the number of cases in which the prediction of the system does not change regardless of the context, i.e., also for the \textit{Neutral} context the prediction of the system remains the same.
This observation confirms our initial hypothesis about the phenomena: the more common entities not only overshadow the less common ones but they are also used as the default predictions made completely independent of the given context, which we call the system priors.
% and (2) prior bias, when the same prediction is made for the neutral context as well, i.e., the prediction of the system does not change regardless of the context. The second case is a subset of overshadowing.
% \todo{is the first case the subset of the second case or do they largely overlap?}

Figure~\ref{fig:prior_bias} shows that among the four best EL systems, REL is the most prone to overshadowing and prior bias.
This also explains its poor performance on the $Shadow$ subset in comparison with the high performance demonstrated on $Tail$.
AIDA and WAT appear to be more sensitive to the local context, which allows them to achieve better results on the overshadowed entities in comparison to both GENRE and REL. Moreover, AIDA, which outperforms all other systems on the \textit{Shadow} subset, turns out to be the least affected by the overshadowing phenomena. These results indicate that the main reason behind the poor ED performance on overshadowed entities is due to systems overrelying on the prior bias and failing to incorporate contextual information.

\begin{figure*}[!htb]
    \begin{subfigure}{0.32\textwidth}
      \includegraphics[width=\linewidth]{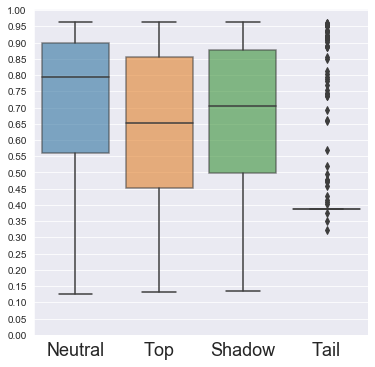}
      \caption{REL}\label{fig:REL_conf}
    \end{subfigure}\hfill
    \begin{subfigure}{0.32\textwidth}
      \includegraphics[width=\linewidth]{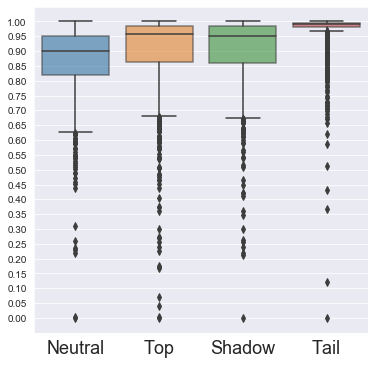}
      \caption{GENRE}\label{fig:GENRE_conf}
    \end{subfigure}\hfill
    \begin{subfigure}{0.32\textwidth}%
      \includegraphics[width=\linewidth]{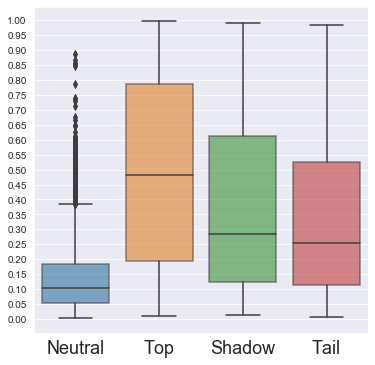}
      \caption{WAT}\label{fig:WAT_conf}
    \end{subfigure}
     \caption{\label{fig:conf_boxplot} Distribution of confidence scores on all subsets of ShadowLink.}
\end{figure*}

Lastly, we also look at the confidence scores for each of the subsets to check if they can be used as an additional indicator (see Figure~\ref{fig:conf_boxplot}).
% \footnote{The outputs of AIDA do not contain its confidence scores.}.
% \todo{what does it show}
Interestingly, the systems have very different distributions of their confidence scores. For example, WAT has lower confidence when given neutral samples, which can be used to detect context ambiguity and filter out such samples.
However, this approach can not be used for REL's and GENRE's predictions\footnote{GENRE's confidence scores were rescaled before the comparison.}.
% , while REL confidently predicts even when there
% \textbf{RQ4:} \textit{How does human performance on overshadowed entities compare with automated results?}
% The results of the manual evaluation (Section~\ref{sec:manual}) show that unlike automated ED systems, human annotators do not appear to rely on prior biases, which allows them to achieve higher scores on overshadowed entities.

% \input{5_results.tex}
\begin{table*}[h!]
\centering
% \resizebox{\columnwidth}{!}{
\begin{tabular}{@{}lcccc@{}}
\toprule
\multicolumn{1}{c}{\textbf{Aspect}} & \textbf{ShadowLink} & \textbf{WikiDisamb30} & \textbf{KORE50} \\ \midrule
\begin{tabular}[c]{@{}l@{}}Source \end{tabular} & Web & Wikipedia & Manual \\
Long-tail entities & \checkmark & - & \checkmark \\
\# Mentions & 15K & 1.4M & 148 \\ \bottomrule
\end{tabular}
% }
\caption{\label{tbl:datasets} ShadowLink in comparison with other datasets that specifically focus on the entity disambiguation task.}
\end{table*}

% \sv{reference \textbf{LOWNER}~\cite{menggemnet} wrt motivating}

\section{Related Work} \todo{make it more structured? or not really?}
% \todo{briefly describe relation between the subsections here}
\textbf{Datasets for ED evaluation.} Evaluation of ED performance was on the research radar for several years, and many benchmark datasets were proposed to date~\cite{hachey2013evaluating,roder2018gerbil,DBLP:conf/clef/EhrmannRFC20}.
Among the most popular ones are AIDA-CONLL~\citep{hoffart2011robust}, which consists of 1.4K annotated news articles with 27.8 entity mentions; AQUAINT dataset~\citep{milne2008learning} with 50 news articles and 727 mentions; MSNBC~\cite{cucerzan2007large} with 20 news articles and 656 mentions.
% \sv{and mentions? if no info we should remove the info on mentions for other datasets}
However, the standard benchmarks used for ED evaluation do not reflect the challenges that are often encountered in practice, such as limited context, long-tail, emerging and complex entities~\cite{menggemnet}.

\citet{guo2018robust} construct two datasets by sampling hard ED examples from Wikipedia and ClueWeb corpora on which a simple baseline using priors does not succeed.
% These datasets were then used to evaluate other ED models.
Their experiments show that this prior-based baseline achieves a high performance, which also indicates the need for more challenging evaluation datasets.
ShadowLink aims to close this gap.
In this work, we focus specifically on the long-tail entities since the existing benchmarks are known to be biased towards the head of the distribution, i.e., the popular entities~\cite{ilievski-etal-2018-systematic,guo2018robust}.

Similarly to ShadowLink, WikiDisamb30~\cite{ferragina2011fast} contains short text snippets annotated with Wikipedia entities designed for ED evaluation.
In contrast to WikiDisamb30, the text snippets in ShadowLink were extracted from web pages outside of Wikipedia to avoid the effects of overfitting since Wikipedia is often used for training language models.
Moreover, ShadowLink examples were collected using Wikipedia disambiguation pages as entity spaces while WikiDisamb30 represents a random sample from Wikipedia that does not allow to examine the effect of overshadowing.

The idea of entity spaces was previously introduced by \citet{van-erp-groth-2020-towards}, who showed that predicting entity spaces largely improves recall.
Their results also hint on the conclusion that disambiguation within entity spaces constitutes a bottleneck in the ED performance.
We take this idea further by designing a dataset centered around entity spaces to evaluate ED performance within entity spaces directly.
This dataset allows us to measure the gap the state-of-the-art ED systems still have on this task.
% We continue this line of work and fill the gap by introducing a new dataset that specifically focuses on disambiguating entity mentions that belong to such entity spaces.

KORE50~\cite{DBLP:conf/cikm/HoffartSNTW12} was created to evaluate the impact of low commonness and high ambiguity on the ED performance but it contains only 50 hand-crafted sentences with 148 entity mentions including ambiguous mentions and long-tail entities. ShadowLink continues this line of work, providing a considerably larger number of samples that can be used for training and evaluation of ED approaches.
We also introduce a subset of neutral samples designed to uncover the model priors.
Table~\ref{tbl:datasets} summarizes how ShadowLink differs from the previously introduced datasets for entity disambiguation.

% new unbiased datasets to 
% ShadowLink differs from other datasets proposed for the ED task. 

% and paired with negative samples.
% , which structurally resembles ShadowLink – however, the purposes and contents of the two datasets are very different.
% WikiDisamb30 was randomly sampled from Wikipedia to provide training data for TagMe, without balancing the data based on commonness, and therefore is not designed for evaluating edge cases of the ED task. 

% , and a novel concept of shadow entities.

% \textbf{Entity spaces.}

% This dataset allows to demonstrate the deficiencies of existing EL systems and provides a new challenging benchmark.

% \todo{change subtitle + check if it looks alright}
\textbf{Robustness evaluation.}  Our approach to ED evaluation taps into the fast-growing area of research aimed at assessing model robustness especially relevant for data-driven machine learning techniques.
One of the first studies on this topic~\cite{sturm2014horse} argued that the state-of-the art music information retrieval systems show very good performance on the standard benchmarks without the real understanding of the task at hand since their predictions relied solely on the confounds present in the ground truth. \citet{sturm2014horse} also coined the term for this phenomena: the "Clever Hans" effect, named after the infamous horse that appeared to solve arithmetic problems while only following unintentional body language cues given by the trainer.
More recently, \citet{lapuschkin2019unmasking} showed that the same effect is demonstrated by other state-of-the-art machine learning models, and the standard performance evaluation metrics fail to detect it. \citet{kauffmann2020clever} further explored this phenomenon, showing that it also affects the reliability of unsupervised models in the field of anomaly detection.
Therefore, not surprisingly we also observed this effect in the ED task: \citet{guo2018robust} used a rudimentary system that merely learned the prior distribution of entities to disambiguate them, and demonstrated that it performs on par with state-of-the-art approaches. These findings specifically calls for new datasets that allow for a more robust evaluation and deeper analysis of the model performance, similar to the one demonstrated here with ShadowLink.
% Comments: All reviewers were in agreement that this paper is clearly written, its arguments well supported, and that the problem of "shadow" entities is important. The presented test set was deemed to be a useful contribution to the entity disambiguation community, and the accompanying analysis contained informative comparisons between baselines as well as between the subsets of "shadow", neutral, and tail entities. Reviewer 3 also suggested that
We hope that this paper might inspire similar datasets in other fields, where the priors from large public datasets may also overshadow the local context.

% \sv{add more related work on robustness evaluation here}
% done

\section{Conclusion}
We introduced ShadowLink, a new benchmark dataset for evaluating entity disambiguation performance, and used it for an extensive analysis of the state-of-the-art systems' results.
Our experimental results indicate that all systems under evaluation are prone to rely on their priors, which explains their higher performance on more common entities, and much lower performance on the lexically similar overshadowed entities.
% All of the EL systems considered in our study have much lower accuracy on the overshadowed entities compared to the top entities that have higher commonness scores.
% Our experiments show that overshadowing impacts the performance of entity disambiguation systems, leading to significantly decreased accuracy. Analysing the prior answers of the models shows that such a performance drop is mainly caused by popularity bias outweighing the impact of local context on the decisions of the models. 
% \todo{what we did? what we learned? what are the implications? what should be done next?}
% Thereby, ShadowLink is designed to stimulate further research on the ED task by showing that it is still far from solved. 
Our work thereby shows that the ED task is still far from solved for overshadowed entities, and ShadowLink paves the way for further research in this direction.

% The most well-known framework for entity linking evaluation is GERBIL \citep{roder2018gerbil}, which provides a consistent way to benchmark entity recognition and linking systems on different datasets.
% \textbf{Future work.}
% First, we intend to integrate the ShadowLink dataset into the GERBIL framework~\citep{roder2018gerbil} to make it easier to use it for experimental evaluation. Second,
The shortcomings of existing disambiguation approaches uncovered by the ShadowLink dataset stimulate further research towards developing more robust ED algorithms that are better at exploiting context without overrelying on the prior bias. We would also like to explore ways to account for more context around the entity mentions, and when expanding the context is actually needed. 
% and better balance the prior belief with the context observation.
% given that we have identified the weaknesses of EL systems towards overshadowed entities due to prior bias, the natural direction of research is 

\section*{Acknowledgements}
This research was supported by
the NWO Innovational Research Incentives Scheme Vidi (016.Vidi.189.039),
the NWO Smart Culture - Big Data / Digital Humanities (314-99-301),
the Informatics Institute of the University of Amsterdam,
the H2020-EU.3.4. - SOCIETAL CHALLENGES - Smart, Green And Integrated Transport (814961),
the Google Faculty Research Awards program.
All content represents the opinion of the authors, which is not necessarily shared or endorsed by their respective employers and/or sponsors.
\todo{added acknowledgements, check if nothing is missing}
\balance
% \newpage
\bibliography{anthology,acl2020}
\bibliographystyle{acl_natbib}
% \newpage
% \appendix

% \input{7_appendix}
\end{document}